\documentclass{IOS-Book-Article}

\usepackage{mathptmx}
\usepackage{soul}\setuldepth{article}
\usepackage{graphicx}
\usepackage{xurl}
\usepackage [english]{babel}
\usepackage [autostyle, english = american]{csquotes}
\MakeOuterQuote{"}
\def\hb{\hbox to 11.5 cm{}}

\begin{document}

\pagestyle{headings}
\def\thepage{}
\begin{frontmatter}              

\title{Weaving Pathways for Justice with GPT}

\subtitle{LLM-driven automated drafting of interactive legal applications}

\markboth{}{December 2023\hb}

\author[A]{\fnms{Quinten} \snm{Steenhuis}\orcid{0009-0001-0110-064X}}
\author[B]{\fnms{David} \snm{Colarusso}\orcid{0009-0003-6287-9284}}
and
\author[A]{\fnms{Bryce} \snm{Willey}\orcid{0000-0003-1775-2869}}

\address[A]{Suffolk University Law School}

\begin{abstract}
Can generative AI help us speed up the authoring of tools to help self-represented litigants?

In this paper, we describe 3 approaches to automating the completion of court forms:
a generative AI approach that uses GPT-3 to iteratively prompt the user to answer
questions, a constrained template-driven approach that uses GPT-4-turbo to generate a draft of
questions that are subject to human review, and a hybrid method. 
We use the open source Docassemble platform in all 3 experiments, together with a tool created at Suffolk University Law School called the Assembly Line Weaver. We conclude that the hybrid model of constrained automated drafting with human review is best suited to the task of authoring guided interviews.
\end{abstract}

\begin{keyword}
document automation \sep generative AI \sep large language models \sep forms \sep form automation
\sep guided interviews \sep interactive legal applications
\end{keyword}
\end{frontmatter}
\markboth{December 2023\hb}{December 2023\hb}

\section{Introduction}
Can generative AI help us speed up the authoring of tools to help self-represented litigants?


The traditional way to help self-represented litigants with court forms is to create a hand-authored interactive legal application, or guided interview. Lauritsen and Steenhuis \cite{lauritsen_substantive_2019} describe a long list of tools that can be used in this traditional approach, with the most popular tools being HotDocs, A2J Author, and Docassemble. A recent entrant in this space focused on legal applications is JusticeBot \cite{westermann_justicebot_2023}.

When we say "forms," we mean the full range of documents that litigators and transactional attorneys might work with. These can include: legal complaints, answers to those complaints, deeds, wills, and demand letters. These documents often look like they only need simple fill in the blanks, but require the application of judgment to be used correctly. For example: wills and trusts have clauses that apply only to some classes of testators. Complaints and answers assert the violation of a law, which may or may not be true depending on individual facts. These rules are not always visible within the four corners of the form.

Creating interactive legal applications with traditional tools is slow and careful work. Authors add markup to templates in PDF or Word format, craft labels for each variable, create pages that place the labeled variables in context, write logical rules to show and hide follow-up questions and text in the final output, and add instructions and help.

Suffolk Law School's Legal Innovation and Technology Lab created a tool to help speed up this task in 2020, during the early
months of the COVID-19 pandemic \cite{steenhuis_digital_2021}. This tool, the Assembly Line Weaver, scans templates for variables and uses a mix of pre-written
questions and heuristics to generate a draft guided interview for the Docassemble \cite{jonathan_pyle_docassemble_2021} platform. The interview itself uses customizable templates for the YAML format that Docassemble uses, and can later be edited by hand.  The tool was designed to allow volunteers to help scale the work of expert form automators, and during the pandemic more than 200 volunteers from around the world helped the lab automate about 30 key processes with it from start to finish \cite{steenhuis_digital_2021}.

Once the hundreds of volunteers left, the lab's work shifted. It didn't make sense for the lab to spend the many hours required to automate each of the almost 800 forms that remained in Massachusetts, even with the Weaver's help. We focused on improving the tools and helping others use them for the most important forms in their own states.

GPT-3 and 4 have inspired computer scientists to experiment with having an AI reorganize legal information. 
Some works have found success in using GPT-4 to convert other forms of legal knowledge, like legislation, into formal structures \cite{2311.04911}. Having an AI do all of the work of forms (scanning a template for variables, asking the user questions, and following up until all variables are defined) is also being debated in hallways at AI and law conferences; touted on Reddit \cite{talktothelampa_how_2023}; and discussed as an idea in general interest magazines \cite{broadway_can_2023}. 
We do not think that it is a good idea except for very simple forms. Instead, we suggest a hybrid approach, with a GPT assisting with the first draft and a human editor turning it into a usable interactive legal application. If it succeeds, this approach can lower the cost to automate court and similar forms dramatically, making it realistic for a state to automate hundreds of forms and reap the benefits for self-represented litigants, including a mobile-friendly interface and integration with electronic filing systems.

Some of our experiments are tentative and early. We acknowledge the limits of our approaches so far. But the promise is hard to deny.

\subsection{Is the form the thing?}
Before we describe our project, we acknowledge the centrality of forms in current approaches to serving self-represented litigants, \textbf{and} that forms may not be the best possible approach to serve that goal.  Branting \cite{branting_narrative-driven_2023} has argued that narratives, not forms, are the tools that lawyers best use to structure effective legal claims. Whether forms or narratives are the best presentation, ultimately, what decision makers need are the facts to make a decision.

Despite that, we start our work with the form. We annotate it, add labels, and turn it into a guided interview in an automated way. There are thousands of existing legal forms (at least 25,000 in the US alone \cite{steenhuis_beyond_2023}) that represent codified legal systems and processes. The work that went into producing the paper form lives in the form's instructions, fields, and checkboxes, ready for the author of an interactive tool to interpret. Even if we want to abandon them eventually, forms are a great place to start. 

\section{Traditional interactive legal application authoring approaches}

An interactive legal application that completes a form has three components:

\begin{enumerate}
    \item An output \textbf{document}, turned into a template with fields. 
    \item A series of \textbf{questions}, organized into one or more screens.
    \item A set of \textbf{logical rules} that decide which followup questions are needed and which boxes or paragraphs are included in the final output.
\end{enumerate}

Human authors perform 4 tasks related to these components:

\begin{enumerate}
    \item \textbf{Label} fields in the template.
    \item Create a brief \textbf{prompt} for each field.
    \item \textbf{Group} fields into a logical order.
    \item Add any conditional logic needed.
\end{enumerate}

This work can take \textbf{hundreds of hours}. One representative project, Massachusetts Defense for Eviction (MADE) \cite{steenhuis_making_2019}, which took about a year to complete, contains:

\begin{enumerate}
    \item 1,100 lines of Python code
    \item 6,141 lines of YAML (or about 30,000 words)
    \item 13 documents in Microsoft Word format, totalling 13,000 words
\end{enumerate}

\subsection{Interactive legal applications require thoughtful choices that affect usability}
While it is tempting to think of the process of creating an interactive legal application from a form as simply letting the user "fill in the blanks," the process requires judgment in several areas. Jarrett and Gaffney's text Forms That Work \cite{jarrett_forms_2008} discusses some of those choices, including: the order of questions, how to provide help, and how to respect the user of your form. The UK's National Health Service \cite{nhs_write_questions_nodate} and SurveyMonkey \cite{survey_monkey_smart_2008} have similar guidance. Our lab developed a detailed style guide with guidance for the authors of legal apps \cite{suffolk_writing_questions_nodate}. We also describe these choices that affect burden in our paper RateMyPDF \cite{steenhuis_beyond_2023}.

Some of these choices can be made once and repeated again and again. For example: we ran usability tests with our questions that ask for a participant's name. We researched and applied best practices to gather a litigant's pronouns and gender. Through practice, we determined an optimal structure for a guided interview that completes a court form. We built the Assembly Line Weaver tool to help us implement these and other choices consistently across each form that we turned into an interactive legal application.

\section{The Assembly Line Weaver}
\begin{figure}
    \centering
    \includegraphics[width=0.5\linewidth]{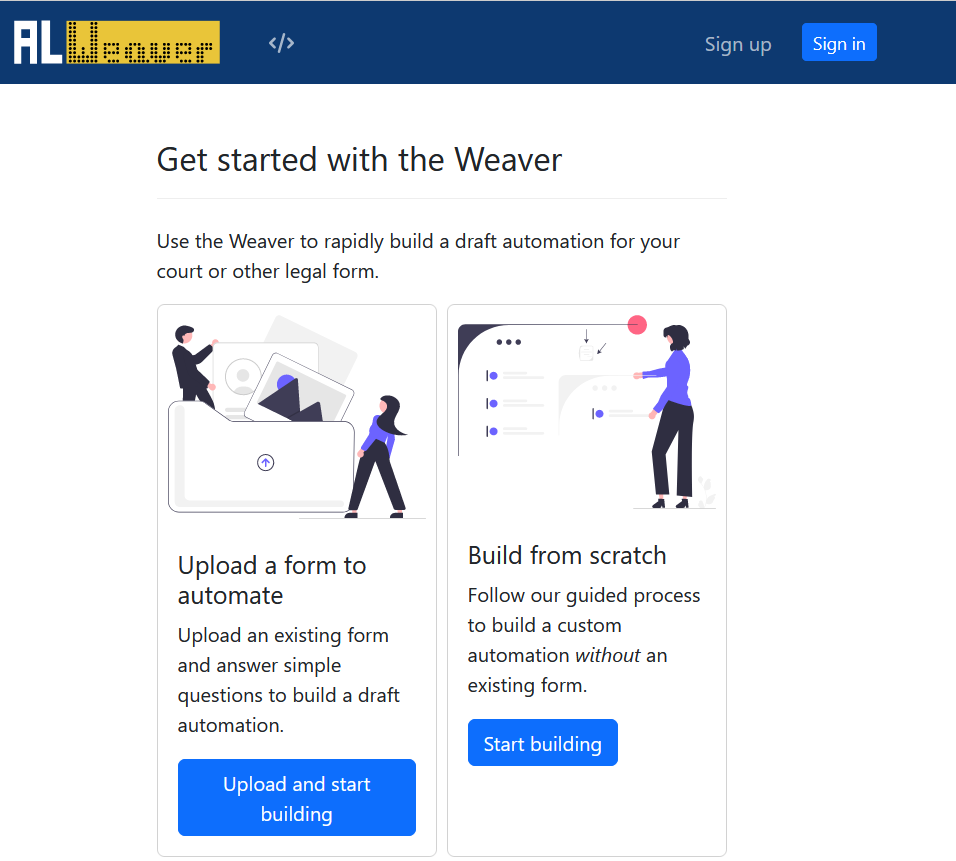}
    \caption{The opening screen of the Assembly Line Weaver}
    \label{fig:weaver}
\end{figure}

The Assembly Line Weaver (Figure \ref{fig:weaver}) is a template-driven guided interview that produces drafts of guided interviews
in the YAML format. It strikes a balance between the ease of use of fully graphical authoring environments and the power and flexibility of Docassemble.

The Weaver:

\begin{enumerate}
    \item Scans a pre-prepared Word or PDF template for field names.
    \item Asks for metadata about the form.
    \item Allows the author to customize the location, branding, and template used to author the interview.
    \item Asks the author to provide "before and after" context to the form, including next steps the user will need to take.
    \item Allows the author to add a prompt for each field.
    \item Asks the author to assign fields to screens, with any extra context that is needed.
\end{enumerate}

In the context of the Weaver, "template-driven" means that we have created a YAML file that has the basic structure of a Docassemble interview. The provided template includes the following screens:
\begin{enumerate}
    \item A title screen.
    \item A "before you start" screen.
    \item Questions, both those created by the author and from the question library for information like party names, pronouns, addresses, and more.
    \item A preview screen.
    \item A "review" screen that allows the user to edit their responses.
    \item A signature screen (with ability to text or email the document for signature by a third party).
    \item A download screen.

\end{enumerate}

The output of the Weaver is a .ZIP file that includes the customized YAML file, the output template, and a template representing the "Next steps" document that can all be further customized. The Weaver does not assist the author with adding logic (or rules) to the guided interview that they produce.

\section{A vision for reducing human effort in form automation}

At our lab, we were interested to see if generative AI could help reduce the human labor needed to create good guided interviews. Ultimately, we prototyped the automatic adding of labels to fields, drafting of prompts, and grouping questions with the assistance of GPT-3 and GPT-4 turbo. We experimented with using ChatGPT+ to convert laws into Python code, but because identifying the statute often requires research outside of the form, we do not plan to integrate this with our automated authoring approach.

In order to maximize our team's productivity, we took two approaches to two unique document types: with Word documents, we focused on identifying and labeling the placeholder variables, which was a significant gap in our existing Weaver tool's capabilities. We left room for these labels to be fed directly into the Weaver in the future, which would take advantage of the structure of our template-driven guided interview authoring. With PDF documents, we prototyped a solution that directly authors a limited version of a Docassemble interview in YAML format. This let us quickly experiment with the interview flow that a large language model would produce. While this still allows for human editing and review, it doesn't take advantage of the existing work we have done with the Weaver to standardize the output. Ultimately, we expect to combine the two approaches.

\section{Auto-labeling of Word documents}

PDF and Word (DOCX) documents have some key differences: PDFs are usually final documents, meant to be printed and filled in by hand at least some of the time. That means that they almost always include lines, boxes, and circles where the user's answer goes. These features can be detected with traditional computer vision tools like OpenCV \cite{bradski_opencv_2000}.

In contrast, Word documents can be edited in a word processor before they are printed and filed. Word documents are more likely to include placeholder text, like [CLIENT NAME], than blank spaces marked by lines or boxes. While one document might consistently use square brackets, another might mix in curly brackets, or leave out the placeholder marker altogether in favor of in-line text like "Your name" or "Dear Merchant." This means that totally rules-based systems, like regular expressions, do not perform well at identifying and labeling placeholders. Statistical approaches, enabled by large language models, can better handle the many variations.

Separately, it was important that whatever method we adopted for labeling the Word document did not disrupt the existing formatting of the document. These features include paragraphs, automatic numbering, and bold and italics, features that are not present in plain Unicode text. That meant the naive approach of sending just the text of the document wouldn't work. Internally, a Word document is an XML document, which \textbf{can} be represented in plain text. However: sending the full uncompressed XML of the document would almost always exceed GPT-4's context window of 4,096 tokens. Although GPT-4 turbo expanded the context window for inputs, the context window for replies remains 4,096 tokens.

We did not want totally free-form labels for variables in our documents. We wanted the automatic labeling to follow some basic rules:

\begin{itemize}
    \item Variables have to be valid Python identifiers, and follow Python conventions like using underscores and lower case letters.
    \item We have a pre-defined set of nouns, like \textbf{users}, \textbf{other\_parties}, \textbf{attorneys}, and more that we preferred to use when applicable, and we always use these nouns as members of a list.
    \item We have a set of predefined attributes for a person, like \textbf{name}, \textbf{address}, and so on that we wanted to use exactly the same way whenever collecting names, addresses, or similar common attributes.
\end{itemize}

We include these variable naming conventions in our prompt to the model.

A Python notebook with our code can be found in our GitHub repository: \url{https://github.com/SuffolkLITLab/FormFyxer/blob/docx-fields-experiment/explore\_labeling\_docx.ipynb}

\subsection{Tokenizing the input document}
To label our Word document, first we use the open source python-docx \cite{noauthor_python-docx_nodate} library to extract the paragraphs and \textbf{runs} of formatted text in each paragraph. These indices in the Python representation can be referenced and used to modify the formatted document later.

Next, we turn the paragraphs and runs of text into a JSON structure, like this:
\begin{verbatim}
    [
        [0, 1, "Dear John Smith:"],
        [1, 0, "I am writing to claim ..."],
        [2, 0, "[Optional: if ...]"],
    ]
\end{verbatim}
Where each entry in the JSON list is a 3-tuple of paragraph number, run number, and the text of that run. Note: this approach does not yet handle tables, which sometimes appeared in our dataset.

\subsection{Prompting the large language model}

\begin{figure}
    \centering
    \includegraphics[width=0.75\linewidth]{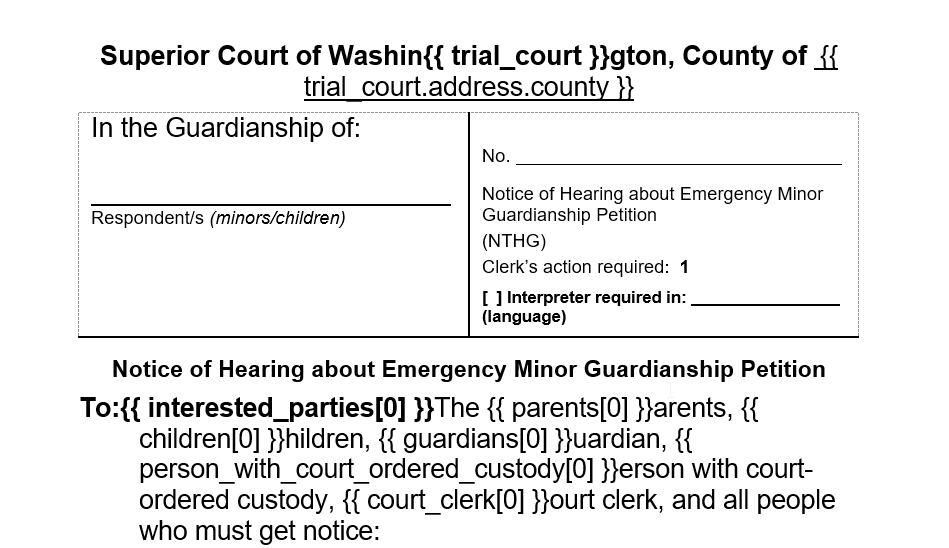}
    \caption{Example of failed GPT-3.5 prompt, with Jinja2 variables inserted in the middle of placeholder phrases}
    \label{fig:failed_prompt}
\end{figure}

Our first attempt was to instruct the model to reply with a JSON structure that includes a tuple of the paragraph number, run number, the Jinja2 variable, and the starting and ending position of the text to be replaced. We also instructed the model to reply with only the modified text, rather than returning the full modified document. In GPT-3.5-turbo, this prompt ignored our instruction to return only the modified runs, instead returning all of the text of the modified document in the reply. In both GPT-3.5 and GPT-4, this prompt failed to accurately mark the start and end position of the text in a way that allowed us to insert Jinja2 variables in context. When we used the positions that were returned by the model, the Jinja2 variables would be inserted haphazardly in the middle of the existing placeholder text, leaving a messy output document that required a lot of manual cleanup. See Figure \ref{fig:failed_prompt}.

A more successful approach (that also used more input tokens) was to prompt the model to return the full text of each modified run. We then iterated through the return value to replace each run with its modified text. Because a run in the DOCX format all has the same formatting, we could safely replace the text of the original run with the modified run. This worked even with very idiosyncratic markup in the example document, as shown below.

\begin{figure}
    \centering
    \includegraphics[width=0.5\linewidth]{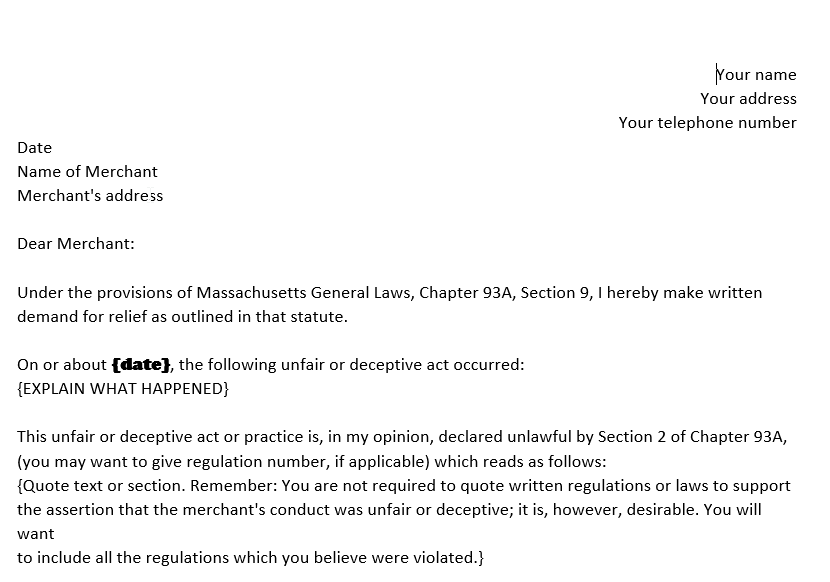}
    \caption{An example document from the wild, demonstrating a wide variety of strategies to mark placeholder text within a single document}
    \label{fig:unlabeled_93a}
\end{figure}
\begin{figure}
    \centering
    \includegraphics[width=0.5\linewidth]{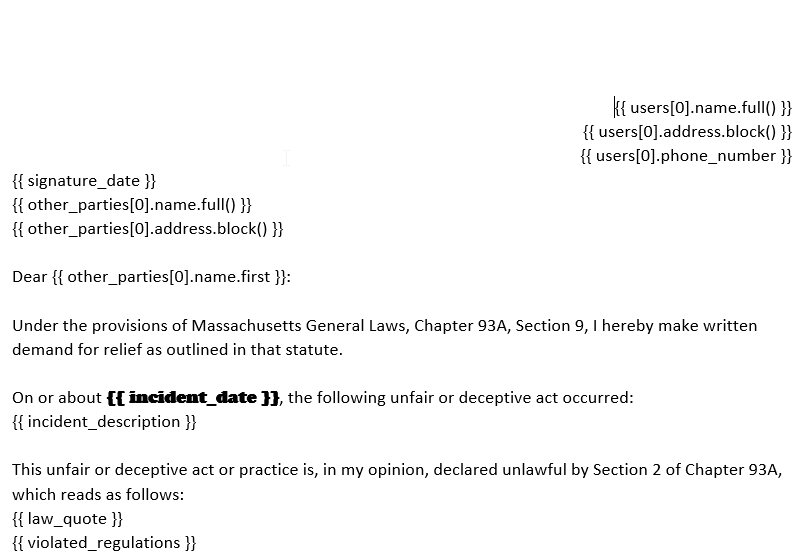}
    \caption{Example of successful output with the revised prompt}
    \label{fig:success_93a}
    
\end{figure}

Our revised prompt succeeding in identifying the placeholder text and replacing it in full. See Figure \ref{fig:success_93a}.

This approach appeared to perform well across a range of documents, a sample of which can be viewed in the GitHub repository. We did not, however, have a large set of pre-annotated documents to benchmark this performance. 

\subsection{Using the output to build an interactive legal application}
Once the Word document is labeled, it can be uploaded to the Weaver. While the author can then assign questions manually for each field, the author can also choose to use our "auto drafting mode," which leverages heuristics and a traditional machine learning approach, but requires a lot of human editing. We describe this existing approach in our paper describing RateMyPDF.\cite{steenhuis_beyond_2023}

\section{Building interactive legal apps automatically with PDF documents}

\begin{figure}
    \centering
    \includegraphics[width=0.5\linewidth]{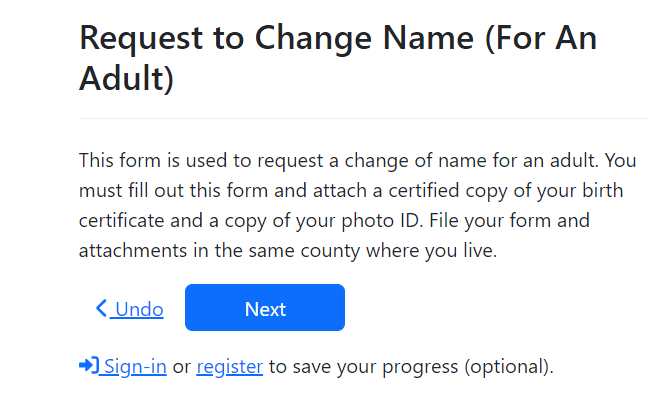}
    \caption{The first page of an automatically generated name change interview}
    \label{fig:name_change_introl}
\end{figure}

\begin{figure}
    \centering
    \includegraphics[width=0.75\linewidth]{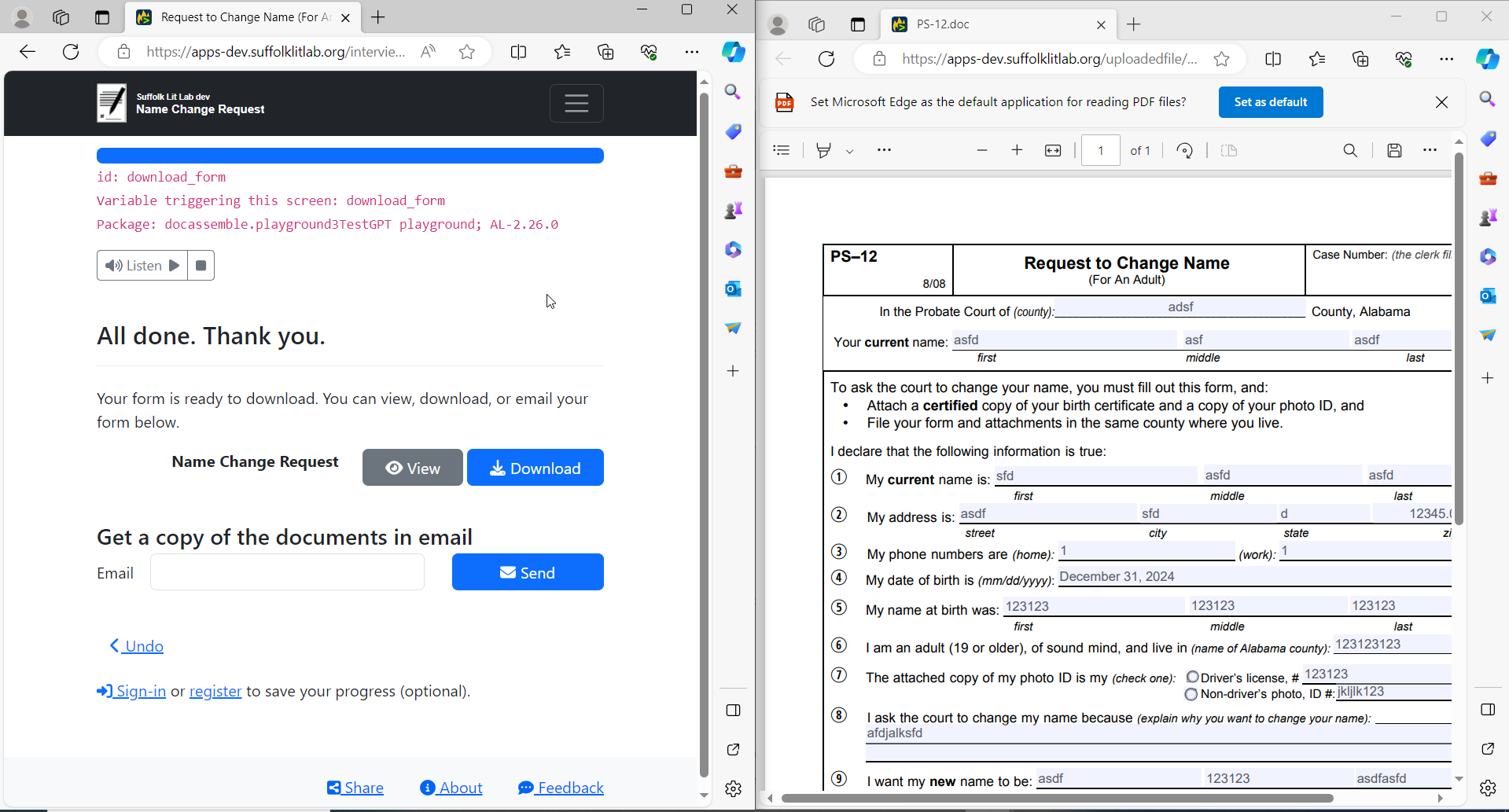}
    \caption{Final screen of the automatically generated name change interview, side by side with the completed PDF}
    \label{fig:output_pdf}
\end{figure}

Our experiment with PDF documents authors Docassemble interviews in YAML format without any human intervention. With important limits, our approach was successful. We generated contextually appropriate labels, questions, and created screens for each question. (Figure \ref{fig:name_change_introl}) The interviews ran and filled in fields in the PDF document. (Figure \ref{fig:output_pdf}) But the PDF's stream-based format made it difficult to accurately identify and label all fields, particularly checkboxes. We describe our success rate and the different ways we worked around this limit below.

We tested this approach with name change forms taken from 12 jurisdictions. We were able to automatically recognize and write questions for 62\% to 69\% of the existing fields. On the best performing form, we were able to do so for 93\% of the fields, compared to 27\% on the worst performing form. For the fields that we could not identify, the problem stemmed from difficulties placing them in context as part of the text provided to the GPT.

All of the code, along with our input and output documents, for this section can be found in an interactive Python notebook, presented here: \url{https://colab.research.google.com/drive/1HRv5imbtHIVYt37h4liwFkW9YlPUNNKI?usp=sharing}.

\subsection{Finding the fields in context} \label{finding_context}

We wanted to feed the GPT model the full text of the form, with the field's placeholders in the right spot in the text, so that the model could come up with appropriate names, just like we did for DOCX files. Ultimately, we found an approach that succeeded most of the time, and we conclude this is a potential place for either further engineering work or adding a human in the loop.

First, we experimented with the native PDF feature of ChatGPT+. As of this writing, you can directly upload a PDF to ChatGPT+ and interact with its contents. We were able to ask ChatGPT+ what fields it thought a PDF of a court form had and get back an answer. However, this feature interacts only with the text, and not the form layer of the PDF. In some cases, it identified fields in the PDF that were not in the form layer, because they were designed to be completed by the court clerk (for example, a "case number" field). In addition, it cannot be used to add text to the PDF at all.

Next, we experimented with text-only approaches. It is possible to get the full text of a PDF with existing open source tools, but this text does not contain any representation of the form layer and its fields. The PDF file format is stream-based, with field elements occupying a stream independent of the PDF's text. Each field element has a defined position on the page, and so does the text, but understanding the relative position of the two requires rendering the PDF, particularly if, as is often the case, there is one large text object in the PDF, with only whitespace added to make room for the PDF fields.

The final, complex, approach that we found was the most reliable was to: 
\begin{enumerate}
    \item fill each field in the form layer with placeholder text (e.g, {{field\_01}})
    \item convert each page of the PDF into a PNG image
    \item apply optical character recognition (OCR) to the images to get the placeholders in context (e.g., "Your name: {{field\_01}}"). 
\end{enumerate}

This approach worked well for fields where the user was expected to write an answer, but it often failed for checkboxes. They were simply too small to add OCR-visible text to. To address this, we did not label small fields in step 1. We reintroduced checkboxes and other small fields when asking the large language model to generate questions, described below. 

\subsection{Leveraging context to create variable names, definitions, and questions}

Given a string for the form's text with variable placeholders in-line, it is possible to have a large language model create appropriate variable names, along with definitions and questions. Building upon the output of step 3 from Section \ref{finding_context}, we can:

\begin{enumerate}
    \item use an LLM to generate a name and description for the document; 
    \item  use an LLM to change placeholder names into semantically appropriate names (e.g., first\_name) based on the output string; 
    \item  use an LLM to write definitions for said variables, again based on the above string; 
    \item  leverage these definitions and an LLM to author questions for users and guess at their data types; and 
    \item  convert these questions into a Docassemble interview by writing a YAML file that takes user answers and fills them into a processed version of the original PDF. 
\end{enumerate}

Checkboxes are handled specially by our prompt. Because we skip checkboxes at our form filling step, at this stage we prompt the large language model to try to assign each checkbox a label and definition based on any text element adjacent to the field on the PDF. This approach could only pair checkbox fields with related text 28\% of the time.

We note that checkbox fields can be important on court forms. The user can ask for relief, identify a legal claim in a complaint, or exercise their right to a jury trial by checking the appropriate box. Therefore, while the OCR method we describe has clear promise, an important future task is to improve our accuracy with checkbox fields.

\subsection{Validating user interactions}

We explored two methods to add input validation in our generation of interviews for PDF forms: a "large language model in the loop" when the user interacted with the form, and an approach that asked the model to classify the input's data type when it first parsed the form.

The "large language model in the loop" approach, while approaching a natural conversation, was resource intensive. After every response by a user, we asked GPT if the answer responded to the question. If not, the user was told why and asked to answer again.  In addition to being expensive, it also risked annoying users by asking too many questions. In testing, it handled redirecting answers like "how is the weather today?" when presenting the user with the question "What is your full name?" well. But we were concerned that the model could be too rigid. For example, would it refuse valid but unusual address lines? What about unusual names, like X? Given the risk of offense and the cost, we abandoned this approach early on.

The up-front classification of answers into datatypes also had risks. For example, phone numbers were often assigned a numeric data type, which precluded users from entering phone numbers of the form 555-5555, accepting only 5555555. ZIP codes were assigned a numeric type too, which strips leading 0's from ZIP codes, causing Massachusetts's ZIP codes to appear incorrect when output. While human review could catch these validation failures, using standardized questions wherever possible, like the Weaver, would cut down the amount of review an author needed to perform.

\subsection{Additional Limitations \& Potential Points of Intervention}

The section above does not address the handling of fields that were neither in the collection of potential checkboxes nor those for which a new name and definition could be made. Only 14\% of the fields in our sample of forms couldn't be placed in-line with the OCR method or identified as a potential checkbox. On the best performing form, every field was either placed in-line or identified as a check box. On the worst performing form, 28\% of the fields remained unidentified. 

The authors of Docassemble interviews can examine and edit the output of the final YAML file to correct any errors and improve the usability of the generated interview.  But in the best-case scenario the full document can be completed. This was most likely to be true when the original PDF had few or no checkboxes. 

We noticed that some errors could compound. For example: a missing or incorrectly labeled field would lead to an incorrect question and an incorrect series of screens. Early review, at multiple stages in the automation, could simplify the author's final editing task.

We identified 2 helpful intervention points:
\begin{itemize}
    \item Before the the title, description, variable names and definitions are finalized
    \item Before questions and data types are finalized
\end{itemize}

We prototyped this review process by allowing the author to edit a JSON object in our Python notebook at each stage.

\section{Reaching level 1}
In our paper Digital Curb Cuts \cite{steenhuis_digital_2021} we described a maturity model for guided interviews, with a level 1 form being at least as good as the experience of completing the original PDF, and level 4 representing a highly polished user experience. Many forms, now locked up in PDFs, would benefit simply from the increased usability of a responsive, mobile friendly design that can be integrated with electronic filing. Our experiments with GPT show that achieving this basic level 1 of automation is possible with a large language model and some Python code alone.

\subsection{Weaving the two approaches together}
We had better success with identifying and labeling fields in Word documents than PDFs, despite the more ambiguous markers that the Word documents used to label placeholders (e.g., see Figure \ref{fig:unlabeled_93a}). But we only used the GPT model to draft questions and put them in order with PDF documents. It's likely that the full interview authoring approach that we took with PDF documents would succeed with labeled Word documents. We could also continue to experiment with solutions to the context problem with PDF checkbox fields.

We think the best way to approach both experimental features is to integrate them with our existing Weaver tool. The unconstrained LLM authoring approach we took with the PDF automation would be enhanced with the template-driven and shared-question approach used in the Weaver. This would limit the need for auto-generated definitions and questions to only those field types not in the Weaver's library. We could get the best of both worlds: vetted, usability tested questions for common fields, and good drafts of the unique questions drafted by the large language model.

\section{Conclusion}

Generative AI can identify user inputs and create interview questions for a wide variety of forms. It can work with both DOCX and PDF formats. For the very simplest forms, especially those without very much conditional logic, large language models can help to produce draft automations that require almost no human intervention at all. However, this AI cannot reliably automate the large number of more complex legal documents without human assistance. Human review and editing of these interviews remains essential. Integrating this human review flexibly at different points in the automation process allows humans to intervene as much or as little as needed, allowing humans to adjust to the needs of a wide range of input documents.

The hybrid approach of automated drafting of a traditional guided interview can get the time saving benefits of the large language model without the risks of fully automated question and answer driven form-filling. It has the potential to significantly expand the number and kind of forms that can benefit from document automation. 

\bibliographystyle{vancouver}
\bibliography{weaving_pathways_to_justice} 

\begin{thebibliography}{10}

\bibitem{lauritsen_substantive_2019}
Lauritsen M, Steenhuis Q.
\newblock Substantive {Legal} {Software} {Quality}: {A} {Gathering} {Storm}?
\newblock In: Proceedings of the {Seventeenth} {International} {Conference} on {Artificial} {Intelligence} and {Law}. Montreal QC Canada: ACM; 2019. p. 52-62.
\newblock Available from: \url{https://dl.acm.org/doi/10.1145/3322640.3326706}.

\bibitem{westermann_justicebot_2023}
Westermann H, Benyekhlef K.
\newblock {JusticeBot}: {A} {Methodology} for {Building} {Augmented} {Intelligence} {Tools} for {Laypeople} to {Increase} {Access} to {Justice}.
\newblock In: Proceedings of the {Nineteenth} {International} {Conference} on {Artificial} {Intelligence} and {Law}. {ICAIL} '23. New York, NY, USA: Association for Computing Machinery; 2023. p. 351-60.
\newblock Available from: \url{https://doi.org/10.1145/3594536.3595166}.

\bibitem{steenhuis_digital_2021}
Steenhuis Q, Colarusso D.
\newblock Digital {Curb} {Cuts}: {Towards} an {Open} {Forms} {Ecosystem}.
\newblock Akron Law Review. 2021;54(4):2.
\newblock Available from: \url{https://ideaexchange.uakron.edu/akronlawreview/vol54/iss4/2/}.

\bibitem{jonathan_pyle_docassemble_2021}
Pyle J. Docassemble; 2021.
\newblock Available from: \url{http://docassemble.org/}.

\bibitem{2311.04911}
Janatian S, Westermann H, Tan J, Savelka J, Benyekhlef K. From Text to Structure: Using Large Language Models to Support the Development of Legal Expert Systems; 2023.

\bibitem{talktothelampa_how_2023}
talktothelampa. How to use {LLM} in order to fill an intake form? [Reddit {Post}]; 2023.
\newblock Available from: \url{www.reddit.com/r/LangChain/comments/140jhln/how_to_use_llm_in_order_to_fill_an_intake_form/}.

\bibitem{broadway_can_2023}
Broadway M. Can {ChatGPT} {Fill} {Out} {Forms}? {Yes}, here's how; 2023.
\newblock Available from: \url{https://www.pcguide.com/apps/can-chatgpt-fill-out-forms-yes-heres-how/}.

\bibitem{branting_narrative-driven_2023}
Branting K, McLeod S.
\newblock Narrative-{Driven} {Case} {Elicitation}.
\newblock Workshop on Artificial Intelligence for Access to Justice (AI4AJ 2023). 2023 Jun.

\bibitem{steenhuis_beyond_2023}
Steenhuis Q, Willey B, Colarusso D.
\newblock Beyond {Readability} with {RateMyPDF}: {A} {Combined} {Rule}-based and {Machine} {Learning} {Approach} to {Improving} {Court} {Forms}.
\newblock Proceedings of International Conference on Artificial Intelligence and Law (ICAIL 2023). 2023:287-96.

\bibitem{steenhuis_making_2019}
Steenhuis Q. Making {MADE}: {User}-centered {Design} in {Practice} – {Quinten} {Steenhuis}; 2019.
\newblock Available from: \url{https://www.nonprofittechy.com/2019/05/12/making-made-user-centered-design-in-practice/}.

\bibitem{jarrett_forms_2008}
Jarrett C, Gaffney G, Krug S.
\newblock Forms that {Work}: {Designing} {Web} {Forms} for {Usability}.
\newblock 1st ed. Amsterdam ; Boston: Morgan Kaufmann; 2008.

\bibitem{nhs_write_questions_nodate}
How to write good questions for forms - {NHS} digital service manual;.
\newblock Available from: \url{https://service-manual.nhs.uk}.

\bibitem{survey_monkey_smart_2008}
{Survey Monkey}.
\newblock Smart {Survey} {Design}.
\newblock Survey Monkey; 2008.
\newblock Available from: \url{https://s3.amazonaws.com/SurveyMonkeyFiles/SmartSurvey.pdf}.

\bibitem{suffolk_writing_questions_nodate}
Writing good questions {\textbar} {The} {Document} {Assembly} {Line} {Project};.
\newblock Available from: \url{https://suffolklitlab.org/docassemble-AssemblyLine-documentation/docs/style_guide/question_overview}.

\bibitem{bradski_opencv_2000}
Bradski G.
\newblock The {OpenCV} {Library}.
\newblock Dr Dobb's Journal of Software Tools. 2000.

\bibitem{noauthor_python-docx_nodate}
python-docx — python-docx 1.1.0 documentation;.
\newblock Available from: \url{https://python-docx.readthedocs.io/en/latest/}.

\end{thebibliography}

\end{document}